\documentclass[10pt,twocolumn,letterpaper]{article}

\usepackage{cvpr}              
\usepackage{amsmath}
\usepackage{amsfonts}
\usepackage{amssymb}
\usepackage{multirow}
\usepackage{colortbl}
\usepackage{booktabs}
\usepackage{arydshln}
\usepackage{algorithm}
\usepackage{algorithmic}

\usepackage[most]{tcolorbox} 
\usepackage{xcolor}
\usepackage{setspace}

\definecolor{cvprblue}{rgb}{0.21,0.49,0.74}
\usepackage[pagebackref,breaklinks,colorlinks,allcolors=cvprblue]{hyperref}

\def\paperID{32264} 
\def\confName{CVPR}
\def\confYear{2026}
\def\modelname{CREM}

\title{\modelname: Compression-Driven Representation Enhancement for Multimodal Retrieval and Comprehension}

\author{
    Lihao Liu$^{1}$\thanks{Equal contributions} \quad 
    Yan Wang$^{2}$\footnotemark[1] \quad 
    Biao Yang$^{2}$\footnotemark[1] \quad 
    Da Li$^{2}$\footnotemark[1] \quad 
    Jiangxia Cao$^{2}$ \quad 
    Yuxiao Luo$^{2}$ \quad
    Xiang Chen$^{2}$ \\
    Xiangyu Wu$^{2}$ \quad 
    Wei Yuan$^{2}$ \quad 
    Fan Yang$^{2}$  \quad
    Guiguang Ding$^{1}$ \quad 
    Tingting Gao$^{2}$ \quad 
    Guorui Zhou$^{2}$ \\
    $^{1}$Tsinghua University \quad $^{2}$Kuaishou Technology
}

\begin{document}
\maketitle
\begin{abstract}
Multimodal Large Language Models (MLLMs) have shown remarkable success in comprehension tasks such as visual description and visual question answering. However, their direct application to embedding-based tasks like retrieval remains challenging due to the discrepancy between output formats and optimization objectives. Previous approaches often employ contrastive fine-tuning to adapt MLLMs for retrieval, but at the cost of losing their generative capabilities. We argue that both generative and embedding tasks fundamentally rely on shared cognitive mechanisms, specifically cross-modal representation alignment and contextual comprehension. To this end, we propose \textbf{\modelname}~(\textbf{C}ompression-driven \textbf{R}epresentation \textbf{E}nhanced \textbf{M}odel), with a unified framework that enhances multimodal representations for retrieval while preserving generative ability. Specifically, we introduce a compression-based prompt design with learnable chorus tokens to aggregate multimodal semantics and a compression-driven training strategy that integrates contrastive and generative objectives through compression-aware attention. Extensive experiments demonstrate that \modelname~achieves state-of-the-art retrieval performance on MMEB while maintaining strong generative performance on multiple comprehension benchmarks. Our findings highlight that generative supervision can further improve the representational quality of MLLMs under the proposed compression-driven paradigm.
\end{abstract}

\section{Introduction}
\label{sec:intro}

\begin{figure}[t]
    \centering
    \includegraphics[width=0.95\linewidth]{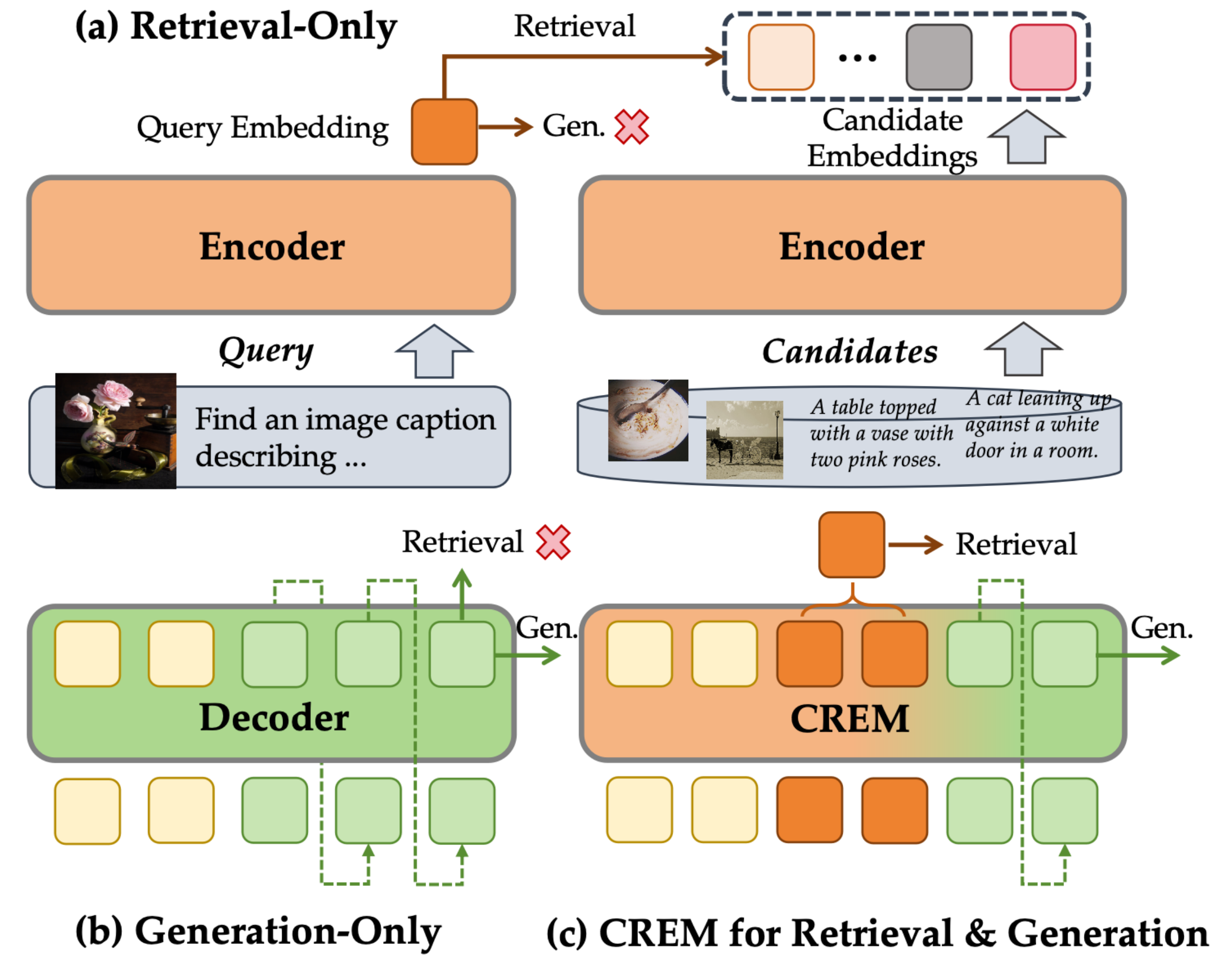}
    \caption{\textbf{Comparison of Different Paradigms. }(a) Embedding models fall short on generation tasks. (b) Generative models lack retrieval capability. (c) Our proposed model \modelname~enables both in a single model.}
    \label{fig:crem-motivation}
\end{figure}

Multimodal Large Language Models (MLLMs) have made significant strides by extending their input capabilities beyond plain text to visual data. These models~\cite{llava,llava1_5,peng2023kosmos,du2021glm,bai2023qwenvl,qwen2vl,internvl2} can integrate information from different modalities and have demonstrated remarkable performance across diverse tasks, including visual question answering, visual grounding, and complex reasoning. A key factor enabling this versatility is that MLLMs can unify these tasks into a conversational format, allowing them to be trained. However, due to the fundamental mismatch between generation and embedding, MLLMs' next-token prediction mechanism limits their capacity to produce high-quality representations for downstream applications such as retrieval and recommendation systems.

Prior studies~\cite{zhang2024magiclens,zhang2024gme,e5v,vlm2vec,unime,liu2025lamra,mmret,chen2025mme5,kong2025unite} have explored transforming MLLMs into embedding models through contrastive learning. These MLLM-based embedding models have demonstrated competitive results, often outperforming traditional CLIP-based models~\cite{clip,blip,jia2021align}. However, after contrastive representation learning, these embedding models lost their original generative capabilities and struggled to complete question-answering tasks, as shown in Fig.~\ref{fig:crem-motivation}(a). This suggests that MLLMs face a trade-off between generation and embedding abilities. Although generation and embedding tasks differ substantially, they both require MLLMs to possess shared capabilities such as cross-modal alignment and contextual reasoning. The relationship between generation and embedding capabilities is like the two sides of a coin. While they share foundational capabilities, optimizing for one often comes at the expense of the other. This phenomenon raises an important question: \textbf{Can MLLMs enhance their representation capability smoothly without compromising generative capability?}

The community has made preliminary explorations. CAFe~\cite{yu2025cafe} introduces a framework that jointly optimizes contrastive and language modeling losses, aiming to unify embedding and generation. Specifically, CAFe utilizes two different prompts to guide MLLM to adapt different tasks (e.g., \textit{i. Compress this image/sentence in one word:} and \textit{ii. Describes the image:}), and establishes a connection between these tasks by simply adding the loss. Actually, such a learning paradigm treats generation and embedding as separate tasks, inevitably leading to suboptimal results: 1) the image/text information must be compressed into limited representation space, and 2) embedding and generation tasks are implicitly modeled independently, ignoring their inherent connection, leading to a trade-off between two tasks.

To overcome these limitations, we introduce \textbf{\modelname}, which is based on a unified framework that leverages learnable \emph{chorus tokens} coupled with a compression-driven training paradigm to seamlessly integrate embedding and generation tasks. The framework is designed to compress visual and textual information into a compact set of special tokens, which serve as a universal representation for diverse downstream applications. The key innovation is a unified prompted-based alignment and a novel compression-aware attention mechanism, which orchestrates feature interactions by constraining chorus tokens to attend to previous input while allowing instructions and answers to focus on the compressed representation. The asymmetric attention design ensures efficient information flow while maintaining task-specific adaptability. Furthermore, we aggregate generation data from different sources to enhance consistency while preserving generalization. With these mechanisms, \modelname~achieves enhanced representation and strong performance without compromising.

We evaluated the performance of \modelname~on the retrieval benchmark MMEB~\cite{vlm2vec} and various comprehension benchmarks, such as MMB~\cite{liu2024mmbench}, MMMU~\cite{yue2024mmmu}. Benefiting from the compression-driven representation enhancement, \modelname~achieves state-of-the-art performance in multimodal retrieval, outperforming the embedding models trained solely on retrieval data by a significant margin, while preserving its language generation capabilities with negligible degradation. 
This illustrates the intrinsic relationship between generation and embedding capabilities, where generative supervision can assist MLLMs in improving embedding quality under unified optimization.
Furthermore, to assess the quality of the compression tokens, we conduct the same comprehension tasks only based on the compressed representations. We observe that even after an 80$\times$ token reduction, the model retains 83\% of its response quality, indicating that the compression tokens preserve sufficient information for retrieval and comprehension. This also demonstrates potential applications in reducing the KV cache size and context length in downstream applications.

The main contributions of this work are summarized as follows:
\begin{itemize}
    \item We propose a \emph{compression-based prompt design} that introduces learnable \emph{chorus tokens} as a bridge between embedding and generation. This design enables broad and consistent representation space for high-quality retrieval embeddings and generative tokens.
    \item We develop a \emph{compression-driven training strategy} that jointly optimizes contrastive learning and language modeling within a unified framework. By incorporating a compression-aware attention mechanism and a generation data mixing strategy, the approach enables dynamic cross-task interaction and efficient knowledge sharing between the two paradigms.
    \item Extensive experiments show that \textbf{\modelname}~achieves state-of-the-art performance on the retrieval benchmark MMEB while maintaining comprehension capabilities. We also conducted extensive analyses to demonstrate that generation can effectively improve retrieval performance.
\end{itemize}

\begin{figure*}[t]
    \centering
    \includegraphics[width=1\linewidth]{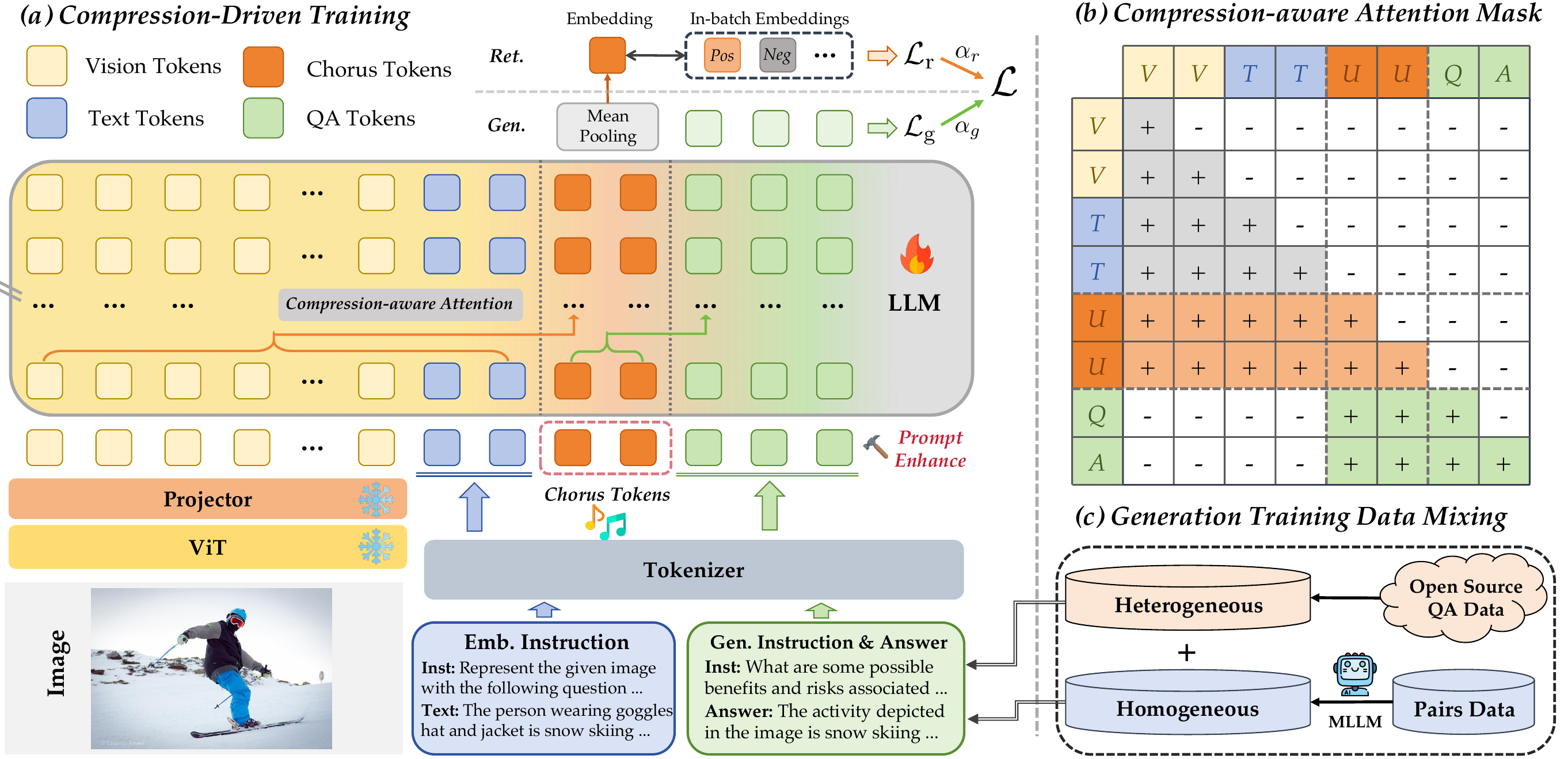}
    \caption{\textbf{Compression-Driven Training Framework of \modelname.}
    (a) The training pipeline integrates chorus tokens with contrastive and generative objectives under a unified prompt design equipped with compression-aware attention. Generation instructions and answers originate from diverse data sources.
    (b) Compression-aware attention mask enforcing token-level visibility constraints, where “+” indicates visible tokens and “–” indicates masked ones.
    (c) Two mixing strategies for generation training using different data sources. Homogeneous data are pseudo-labeled by an MLLM from retrieval pairs, whereas heterogeneous data are collected from open-source datasets.}
    \label{fig:crem-framework}
\end{figure*}

\section{Related Work}
\label{sec:related_work}

\paragraph{Multimodal Large Language Models}
MLLMs extend LLMs by jointly processing and integrating cross-modal information~\cite{peng2023kosmos,zhu2023minigpt,bai2023qwenvl,llava,minicpmv}. Early work like LLaVA~\cite{llava,llava1_5,llavanext} integrates a vision encoder via projection to convert visual inputs into language-compatible tokens. LLaVA-OneVision~\cite{llava_onevision} consolidates LLaVA series from data, model and training strategy. Predominant models such as Qwen-VL~\cite{qwen2vl,qwen2_5vl} and InternVL~\cite{internvl2,internvl3} series further advance multimodal understanding through architectural innovations, improved training, and large-scale datasets, supporting complex tasks.

\vspace{-0.7em}
\paragraph{Multimodal Representation Learning} 
Models such as CLIP~\cite{clip}, SigLIP~\cite{siglip}, and CoCa~\cite{coca} learn aligned representations from weakly supervised image-text pairs, typically by encoding each modality independently. Recent approaches~\cite{e5v,vlm2vec,mmret,mmembed,zhang2024gme,unime,chen2025mme5,lan2025llave,kong2025unite,vlm2vec_v2} aim to leverage the rich pre-training knowledge embedded in MLLMs to construct high-quality universal embeddings. 
For instance, E5-V~\cite{e5v} adopts a unimodal training paradigm that surpasses conventional multimodal methods in image-text retrieval. 
VLM2Vec~\cite{vlm2vec,vlm2vec_v2} introduces a contrastive learning framework capable of processing instruction-based multimodal pairs. 
mmE5~\cite{chen2025mme5} employs a data synthesis strategy to enhance multimodal multilingual embeddings, while UniME~\cite{unime} boosts performance through textual knowledge distillation and hard-negative-aware instruction tuning. 
LLaVE~\cite{lan2025llave} further improves multimodal embeddings by exploiting the discriminative difficulty of negative samples. 
UNITE~\cite{kong2025unite} conducts a systematic analysis of modality-specific data and proposes a modality-aware training scheme to alleviate the competition among cross-modal instances. 
Despite these advances, MLLMs often compromise their generative capacity when adapted for embedding tasks, and the transfer process remains suboptimal, potentially leading to the loss of generative knowledge.

\vspace{-0.8em}
\paragraph{Unified Generative Embedding Models}
Recent studies~\cite{gritlm,sugar,mmgem,ouali2025vladva,yu2025cafe} have explored the unification of generative and embedding objectives within a single framework to overcome the limitations of task-specific architectures. 
GRITLM~\cite{gritlm} employs instruction tuning to enable large language models to flexibly switch between generation and embedding modes. 
Sugar~\cite{sugar} introduces a structure-induced training strategy that jointly models discriminative and generative capabilities through interleaved image-text sequences. 
MM-GEM~\cite{mmgem} demonstrates that a unified vision-language architecture with a shared pooling mechanism can effectively support both tasks, achieving competitive results in retrieval and captioning. 
VladVA~\cite{ouali2025vladva} leverages short and long captions for joint autoregressive learning and image-text contrastive alignment. 
CAFe~\cite{yu2025cafe} integrates contrastive and autoregressive objectives to fine-tune MLLMs on detailed image-text descriptions, enhancing both retrieval accuracy and generative coherence.
However, most of these methods rely on the simple combination of two independent loss functions. 
A truly unified paradigm that simultaneously enhances representation learning from both generative and embedding perspectives remains underexplored.

\vspace{-0.8em}
\paragraph{Multimodal Token Compression.} 
In MLLMs, token compression aims to condense redundant vision tokens into compact representations to alleviate quadratic computational costs and mitigate context overflow during generation. 
A range of compression strategies have been explored. 
Some methods perform token pruning at the LLM input layer~\cite{prumerge,tokenpacker,xu2024slowfast,chu2023mobilevlm}, while others progressively discard tokens across encoder or decoder layers based on attention maps or similarity measures~\cite{fastv,pyramiddrop,yang2025visionzip,zhang2024sparsevlm}. 
While Perceiver Resampler~\cite{perceiver,flamingo,bai2023qwenvl} and Q-Former~\cite{blip2} aggregate dense visual tokens via cross-attention, other works~\cite{victor,vocollama} leverage the self-attention capabilities of LLMs to compress comprehensive visual information. CoMa~\cite{li2025coma} introduces a compressed pre-training phase that explicitly decouples information coverage from discriminative matching. Such advancements underscore our observation that token compression for generative tasks shares an intrinsic objective with embedding learning.
\section{Method}
\label{sec:method}

As illustrated in Fig.~\ref{fig:crem-framework}(a), 
\modelname~introduces a set of learnable \textit{chorus tokens} designed to store and share condensed semantic information that can be jointly utilized for both embedding and generation tasks. 
During training, these special tokens are appended to the original multimodal inputs to perform semantic compression of raw tokens. 
Based on the resulting compressed representation, we jointly optimize two complementary objectives: 
1) for embedding task, we apply contrastive learning on the pooled representation; 
2) for generation task, we constrain the model to produce responses solely from the compressed representation. 
This compression-driven framework enhances the representational capacity for information storage while aligning the optimization of fundamentally different tasks within a consistent representation space. 
In addition, we observe that compressed representations naturally serve as effective substitutes for redundant multimodal tokens during inference, reducing the KV cache size while preserving a substantial portion of the model’s comprehension ability.

\subsection{Compression-Based Prompt Design}
\label{sec:chorus-token-design}

For multimodal generation, the model prompt is typically organized with a structured template~\cite{llava,internvl2,qwen2vl}, where the \textit{system} and \textit{user} provide instructions, and responses are generated after the \textit{assistant} based on preceding tokens.
In contrast, embedding models~\cite{e5v,vlm2vec,unime} often adopt inconsistent prompt templates and use the EOS token as the encoded representation. 
From the retrieval perspective, the EOS token serves as an aggregated representation of the visual content, while from the comprehension perspective, visual feature tokens capture fine-grained spatial and semantic details. 
These two paradigms exhibit distinct characteristics: retrieval progressively distills visual features into compact EOS representation, whereas comprehension relies on sparse interactions among numerous vision tokens. 

Previous works~\cite{bolya2022tome,fastv,yang2025visionzip,zhang2024sparsevlm,vocollama} have demonstrated that the existing visual representation contains substantial spatial and semantic redundancy, suggesting a significant potential for compression. 
Inspired by this, we propose reconstructing retrieval-optimized compressed features (i.e., EOS tokens) as semantic primitives for comprehension tasks. 
This reconstruction forms a unified representational unit, termed the \textit{chorus token}, which serves as a shared vehicle bridging comprehension and retrieval. 
Through this unified representation, our framework harmonizes the dual objectives of retrieval and comprehension in an end-to-end manner, distilling visual information into compact yet semantically rich chorus tokens. 

\begin{tcolorbox}[
    colframe=black, 
    arc=6pt,
    boxrule=0.8pt,
    left=6pt, right=6pt, top=5pt, bottom=5pt,
    enhanced
]

\textbf{System:} \textit{You are a helpful assistant that can represent and understand multimodal inputs.} \\
\textbf{User:} \texttt{<image>} \texttt{[eInst]} \texttt{<chorus>} \texttt{[gInst]} \\
\textbf{Assistant:} \texttt{<answer>}
\end{tcolorbox}

To achieve this, we unify the prompt design for both tasks. Specifically, 
the embedding process is reformulated to follow the generation-style prompt, fully leveraging the inherent instruction follow-up capability of the model. 
Then we insert chorus tokens $\mathcal{U}$ (\texttt{<chorus>}) into the \textit{user} content between the embedding instruction $\mathcal{T}$ (\texttt{[eInst]}) and the generation instruction $\mathcal{Q}$ (\texttt{[gInst]}), serving as a compressed representation of the preceding vision tokens $\mathcal{V}$ (\texttt{<image>}) and textual tokens $\mathcal{T}$. 
During generation, the \textit{assistant} produces responses $\mathcal{A}$ (\texttt{<answer>}) conditioned on the chorus representation, which acts as an efficient surrogate for the full multimodal inputs. 
Hence, both tasks are formulated as a joint optimization objective that maximizes the mutual information between the chorus tokens and the multimodal representations.
\begin{equation}
\mathbb{I}_{ \mathcal{V,T};\mathcal{U}} = D_{\text{KL}} \left( p( \mathcal{V}, \mathcal{T},\mathcal{U}) \parallel p( \mathcal{V}, \mathcal{T}) \otimes  p( \mathcal{U}) \right) 
\end{equation}
As illustrated in Fig.~\ref{fig:crem-framework}(b), the vision tokens $\mathcal{V}$ and text tokens $\mathcal{T}$ are visible only to the chorus tokens $\mathcal{U}$, while remaining hidden from the question tokens $\mathcal{Q}$ and answer tokens $\mathcal{A}$. 
To achieve our goal of seamlessly adapting an existing model into a unified framework, we regulate this visibility through a compression-aware attention mask $M$. 
Specifically, we modify the standard causal attention mask by restricting the attention flow from QA tokens to the original textual and vision tokens, as defined below:
\begin{equation}
M_{i j} = \begin{cases} 
0, & \text{if } i \in(\mathcal{Q},\mathcal{A}) \text{ and } j \in(\mathcal{V}, \mathcal{T}), \\
1(j\leq i), & \text{otherwise}.
\end{cases}
\end{equation}

\begin{figure}[t]
    \centering
    \includegraphics[width=0.95\linewidth]{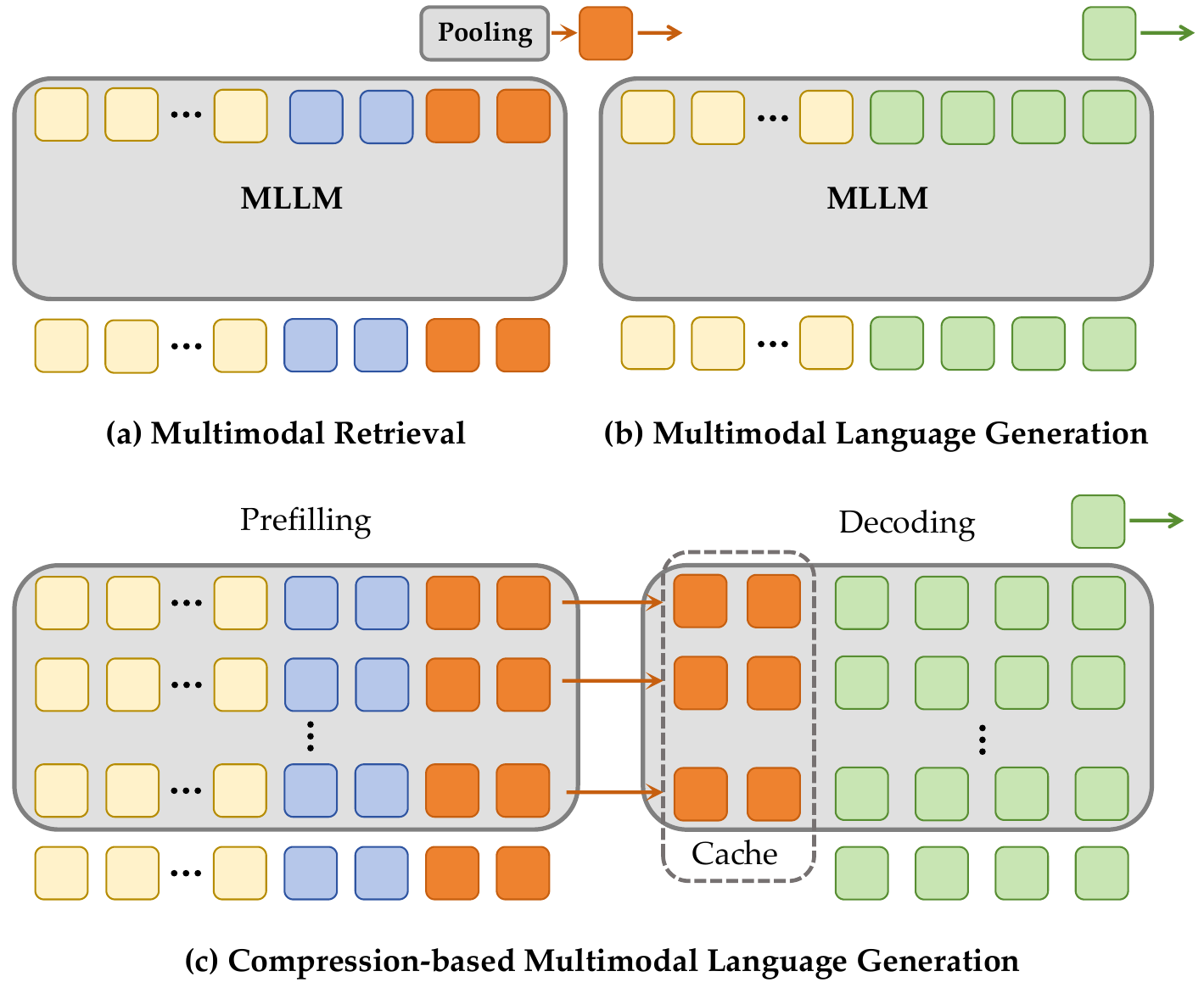}
    \caption{\textbf{\modelname~Inference Modes.} (a) Retrieval embeddings are derived from pooled chorus tokens. (b) Native next-token prediction is performed with full access to all vision tokens (\textit{Nat.}). (c) Efficient inference is achieved by pruning vision tokens and reducing KV caches (\textit{Comp.}). }
    \label{fig:crem-infer}
\end{figure}

\subsection{Compression-Driven Training Strategy}
\label{sec:multitask-train}

Building on the above methodology, we integrate retrieval and generation into a shared optimization space. We further introduce two generation data mixing strategies to refine this space, as illustrated in Fig.~\ref{fig:crem-framework}(c).
\textbf{1) Homogeneous Data Mixing:} both tasks utilize the same samples, where each retrieval pair is augmented with QA-style data generated by an off-the-shelf MLLM. For image–text samples, descriptive answers are produced based on the image and instruction. For text-only data (e.g., captions or labels), \texttt{[gInst]} is set as \textit{``Reconstruct the represented text''} to induce text reconstruction through generation. Both contrastive and generative losses are computed on the same sample to encourage cross-task consistency. 
\textbf{2) Heterogeneous Data Mixing:} retrieval and generation samples are drawn from different sources but optimized jointly within each batch. For generation samples, \texttt{[eInst]} is set as \textit{``Represent the given image.''} and \texttt{[gInst]} corresponds to the image-related query. Through gradient accumulation, the two tasks are mixed within a single batch, where task-specific losses are computed independently and accumulated before a unified backward pass. The inclusion of diverse generative data enhances fine-grained representation learning and helps maintain strong generalization in multimodal comprehension.

For contrastive learning, we follow the instruction-based multimodal contrastive learning framework proposed in~\cite{vlm2vec}. Based on image-text pair data, we construct positive sample pairs by synthesizing instruction-driven queries $q$ and targets $t$, and compute a standard InfoNCE loss over in-batch negatives for retrieval task:
\begin{equation}
\mathcal{L}_{\text{r}} = -\log \frac{\phi(\mathbf{h}_q, \mathbf{h}_{t^+})}{\phi(\mathbf{h}_q, \mathbf{h}_{t^+}) + \sum\limits_{t^- \in \mathbb{N}} \phi(\mathbf{h}_q, \mathbf{h}_{t^-})}
\end{equation}
Here, $\mathbb{N}$ denotes the set of all negatives, $\phi$ is a function that computes the matching score. And we adopt the cosine similarity function as $\phi(\mathbf{h}_q,\mathbf{h}_{t})=\exp(\cos(\mathbf{h}_q, \mathbf{h}_{t})/\tau)$, where $\tau$ is a temperature hyperparameter. The representation $\mathbf{h}_q$ and $\mathbf{h}_t$ are obtained via mean pooling over the representation $\mathbf{h}_u$ of the $k$ learnable chorus tokens.

To facilitate training convergence while preserving the model's native generative capability, we introduce \emph{stochastic compression-driven language modeling loss}. 
Specifically, we define a Bernoulli random variable $z \sim \mathrm{Bernoulli}(p)$, which controls whether the generative model conditions on the full multimodal context ($z=0$) or solely on the compressed representation ($z=1$). 
The generative objective can thus be formalized as:
\begin{equation}
    \mathcal{L}_{\text{g}} = -\frac{1}{T}\sum_{t=1}^T\log p_C(y_t \mid y_{<t},\mathcal{U},1_{z=0}(\mathcal{V},  \mathcal{T})).
\end{equation}
This stochastic formulation encourages the model to maintain both compression robustness and generative fluency.

Thus, the overall multitask objective is a weighted combination of contrastive and language modeling losses:
\begin{equation}
\mathcal{L} = \alpha_{\text{r}} \mathcal{L}_{\text{r}} + \alpha_{\text{g}} \mathcal{L}_{\text{g}}
\end{equation}

\subsection{Multi-Task Inference Modes}
\label{sec:ittc} 

Our compression-driven training not only improves representation quality but also enables efficient generative inference. 
The resulting compressed representations can serve both as multimodal embeddings and as plug-and-play KV caches. 
By discarding the preceding multimodal tokens during decoding, we support longer input contexts and reduce memory consumption.

For the retrieval task, since downstream processing is non-autoregressive and outputs depend solely on preceding tokens, the inference procedure remains largely consistent with conventional pipelines. 
All chorus representations from the final layer are pooled to form a multimodal embedding, as shown in Fig.~\ref{fig:crem-infer}(a).
For the generation task, we adopt a dual-path decoding strategy. 
As illustrated in Fig.~\ref{fig:crem-infer}(b), the model can process multimodal inputs in the native format without the insertion of chorus tokens $\mathcal{U}$. 
In compression-based generation, the chorus tokens $\mathcal{U}$ are employed in a single forward pass to aggregate information, as shown in Fig.~\ref{fig:crem-infer}(c). 
These tokens can then populate the KV cache for decoding or be stored for future reuse, thereby eliminating redundant computation and reducing memory usage in long-context scenarios.

\section{Experiments}
\label{sec:experiments}

\begin{table*}[t]
\centering
\caption{\textbf{Results on MMEB.} ``IND'' denotes in-distribution datasets, while ``OOD'' refers to out-of-distribution datasets. Reported scores are the average Precision@1 across the respective dataset groups. The highest score in each column is shown in bold, and the second-best result is underlined.}
\label{tab:mmeb}
\resizebox{\textwidth}{!}{
\begin{tabular}{lccccccccc} 
\toprule
\multirow{2}{*}{\textbf{Model }} & \multirow{2}{*}{\textbf{Backbone}} & \multirow{2}{*}{\textbf{\#Params}} & \multicolumn{4}{c}{\textbf{Per Meta-Task Score }} & \multicolumn{3}{c}{\textbf{ Average Score}} \\ 
\cmidrule(lr){4-7} \cmidrule(lr){8-10} 
 &  &  & Classification & VQA & Retrieval & Grounding & IND & OOD & Overall \\ 
\midrule
\# of Datasets $\to$ &  &  & 10 & 10 & 12 & 4 & 20 & 16 & 36 \\ 
\midrule
CLIP~\cite{clip} & - & 0.4B & 55.2 & 19.7 & 53.2 & 62.2 & 47.6 & 42.8 & 45.4 \\
OpenCLIP~\cite{cherti2023openclip} & - & 0.4B & 56.0 & 21.9 & 55.4 & 64.1 & 50.5 & 43.1 & 47.2 \\ 
\hline
\multicolumn{10}{c}{{\cellcolor[rgb]{0.91,0.91,0.91}}\textbf{\textit{\textless{} 3B Models }}} \\ 
\hline
VLM2Vec~\cite{vlm2vec} & Qwen2-VL & 2B & 59.0 & 49.4 & 65.4 & 73.4 & 66.0 & 52.6 & \textcolor[rgb]{0.153,0.153,0.165}{59.3} \\
VLM2Vec-V2~\cite{vlm2vec_v2} & Qwen2-VL & 2B & 62.9 & 56.3 & \textbf{69.5} & 77.3 & 68.7 & \underline{60.1} & 64.9 \\
GME~\cite{zhang2024gme} & Qwen2-VL & 2B & 54.4 & 29.9 & 66.9 & 55.5 & - & - & 51.9 \\
UNITE~\cite{kong2025unite} & Qwen2-VL & 2B & \underline{63.2} & 55.9 & 65.4 & 75.6 & 65.8 & \underline{60.1} & 63.3 \\
LLaVE~\cite{lan2025llave} & Aquila-VL & 2B & 62.1 & \underline{60.2} & 65.2 & \textbf{84.9} & 69.4 & 59.8 & \underline{65.2} \\
CAFe~\cite{yu2025cafe} & LLaVA-OV & 1B & 59.1 & 49.1 & 61.0 & \underline{83.0} & 64.3 & 53.7 & 59.6 \\
\rowcolor[rgb]{0.914,0.914,1} \textbf{\modelname~(Ours)} & Qwen2-VL & 2B & \textbf{65.8} & \textbf{60.7} & \underline{68.3} & 78.9 & \textbf{70.8} & \textbf{61.5} & \textbf{66.7} \\ 
\hline
\multicolumn{10}{c}{{\cellcolor[rgb]{0.91,0.91,0.91}}\textbf{\textbf{\textit{\textgreater{} 7B Models}}}} \\ 
\hline
E5-V~\cite{e5v} & LLaVA-1.6 & 7B & 39.7 & 10.8 & 39.4 & 60.2 & 34.2 & 33.9 & 33.9 \\
MMRet~\cite{mmret} & LLaVA-1.6 & 7B & 56.0 & 57.4 & 69.9 & 83.6 & 68.0 & 59.1 & 65.8 \\
VLM2Vec~\cite{vlm2vec} & Qwen2-VL & 7B & 62.6 & 57.8 & 69.9 & 81.7 & 72.2 & 57.8 & 65.8 \\
UniME~\cite{unime} & LLaVA-OV & 7B & 66.8 & \underline{66.6} & 70.5 & 90.9 & - & - & \underline{70.7} \\
UNITE~\cite{kong2025unite} & Qwen2-VL & 7B & \textbf{68.3} & 65.1 & \underline{71.6} & 84.8 & 73.6 & 66.3 & 70.3 \\
mmE5~\cite{chen2025mme5} & Llama-3.2-Vision & 11B & 67.6 & 62.8 & 70.9 & \underline{89.7} & 72.3 & \underline{66.7} & 69.8 \\
CAFe~\cite{yu2025cafe} & LLaVA-OV & 7B & 65.2 & 65.6 & 70.0 & \textbf{91.2} & \textbf{75.8} & 62.4 & 69.8 \\
\rowcolor[rgb]{0.914,0.914,1} \textbf{\textbf{\modelname~(Ours)}} & Qwen2-VL & 7B & \textbf{68.3} & \textbf{69.4} & \textbf{72.9} & 86.1 & \underline{75.6} & \textbf{67.8} & \textbf{72.1} \\
\bottomrule
\end{tabular}}
\end{table*}

\paragraph{Training Datasets} 
Owing to consistent optimization strategy, \modelname~naturally incorporates both retrieval and generation capabilities. 
To activate these functionalities effectively during training, we employ two categories of data: retrieval-oriented and generation-oriented datasets. 
For retrieval training, we use the training split of the Massive Multimodal Embedding Benchmark (MMEB)~\cite{vlm2vec}, which contains datasets from four meta-task categories: classification, visual question answering, retrieval, and visual grounding.
For generation-oriented training, we utilize two complementary sources. 
The heterogeneous ShareGPT-4V dataset~\cite{chen2024sharegpt4v} is employed to enhance token compression while maintaining model generalization. 
In addition, for homogeneous generation data, we use Qwen2.5-VL-7B~\cite{qwen2_5vl} to synthesize QA-style data corresponding to each MMEB sample, enabling consistent cross-task supervision.

\vspace{-0.8em}
\paragraph{Evaluation and Metrics} 
We evaluate the retrieval performance on the MMEB benchmark, which also provides comprehensive evaluation splits. According to the meta-task taxonomy, the 36 evaluation datasets are divided into 20 in-distribution and 16 out-of-distribution subsets. We adopt \textbf{Precision@1} as the primary metric, emphasizing the correctness of top-ranked candidates.
For generative evaluation, we conduct a broad and systematic comparison. To rigorously assess the improvement introduced by our approach, we employ multiple established benchmarks, including MMB~\cite{liu2024mmbench}, MMVet~\cite{yu2023mmvet}, AI2D~\cite{kembhavi2016ai2d}, HallusionBench~\cite{guan2024hallusionbench}, MMMU~\cite{yue2024mmmu}, and MMStar~\cite{chen2024mmstar}, following the evaluation protocol of Qwen-VL~\cite{qwen2vl}. 
We report the averaged performance under two settings: 
1) \textbf{\textit{Nat.}}, all vision tokens are fully attended to assess the model’s intrinsic generative ability; 2) \textbf{\textit{Comp.}}, only the chorus tokens are retained as compressed visual context to evaluate the fine-grained fidelity of the compressed representations.

\vspace{-0.8em}
\paragraph{Implementation Details}
We adopt Qwen2-VL~\cite{qwen2vl} as the backbone model and train it using LoRA with rank 16 and alpha 64. The number of chorus tokens $k$ is set to 16 by default. Each training batch consists of 1024 retrieval-labeled samples with homogeneous generation labels and 128 heterogeneous generation-labeled samples. Following VLM2Vec~\cite{vlm2vec,vlm2vec_v2}, we apply GradCache~\cite{gradcache} to enlarge the effective per-device batch size and adopt the interleaved sampling strategy for retrieval tasks, where a global batch is divided into $n$ sub-batches, each corresponding to a distinct dataset. The model employs dynamic resolution, limiting the number of vision tokens to at most 1280, with a total context length capped at 2048. Training is performed for 2000 steps with a learning rate of $5e^{-5}$ and a 100-step warm-up. The loss weights for retrieval and generation are set to $\alpha_{\text{r}}=1$ and $\alpha_{\text{g}}=0.5$, respectively. The compression probability $p$ in the generation loss is fixed at 0.5.

\begin{table*}[t]
\setlength{\extrarowheight}{1pt}
\setlength{\aboverulesep}{0.5pt}
\setlength{\belowrulesep}{0.5pt}
\centering
\caption{\textbf{Results on Multimodal Comprehension Benchmarks.} The comprehension benchmarks are used to evaluate the multimodal generative capability of different models. 
\modelname$_{G}$ refers to Qwen2-VL fine-tuned solely on ShareGPT-4V datasets using the standard generative pipeline, while \modelname$_{R}$ denotes Qwen2-VL trained exclusively on MMEB following the VLM2Vec~\cite{vlm2vec}. ``AVG'' denotes the average score across all benchmarks. 
All evaluations are conducted using the same image resolution as in the retrieval experiments.}
\label{tab:mllm}
\begin{tabular}{lccccccc} 
\toprule
\textbf{Model} & \textbf{MMB} & \textbf{MMVet} & \textbf{AI2D} & \textbf{Hallusion} & \textbf{MMMU} & \textbf{MMStar} & \textbf{AVG} \\ 
\midrule
\multicolumn{8}{c}{{\cellcolor[rgb]{0.91,0.91,0.91}}\textit{\textbf{2B Models}}} \\ 
\hline
Qwen2-VL & 72.3 & 45.7 & 73.8 & 41.9 & 41.1 & 46.1 & 53.5 \\
\modelname$_G$ & 73.1 & 44.5 & 72.9 & 41.0 & 41.2 & 46.2 & 53.2 \\
\modelname$_R$ & 64.3 & 17.3 & 66.7 & 33.6 & 34.9 & 43.9 & 43.4 \\
\rowcolor[rgb]{0.914,0.914,1} \textbf{\modelname} & 72.5 & 45.1 & 72.8 & 41.2 & 41.4 & 45.5 & 53.1 \\ 
\hline
\multicolumn{8}{c}{{\cellcolor[rgb]{0.91,0.91,0.91}}\textbf{\textit{7B Models}}} \\ 
\hline
Qwen2-VL & 80.9 & 58.0 & 82.2 & 50.9 & 53.7 & 59.5 & 64.2 \\
\modelname$_G$ & 80.7 & 56.8 & 81.9 & 49.2 & 51.7 & 59.8 & 63.4 \\
\modelname$_R$ & 77.3 & 41.4 & 80.9 & 44.5 & 47.0 & 56.9 & 58.0 \\
\rowcolor[rgb]{0.914,0.914,1} \textbf{\modelname} & 80.5 & 56.7 & 81.9 & 48.8 & 52.1 & 59.3 & 63.2 \\
\bottomrule
\end{tabular}
\end{table*}
\section{Main Results}
\label{sec:results}

\paragraph{Multimodal Retrieval}
As shown in Tab. \ref{tab:mmeb}, \modelname~outperforms retrieval-specialized models, even though it adopts only in-batch negative training strategy and the same data schedule as VLM2Vec~\cite{vlm2vec,vlm2vec_v2}. 
\modelname~outperforms UNITE~\cite{kong2025unite}, which is trained on large-scale multi-source data, UniME~\cite{unime}, which employs a hard negative sampling strategy, and larger models trained with synthetic data such as mmE5~\cite{chen2025mme5}. In comparison to CAFe~\cite{yu2025cafe}, \modelname~achieves superior retrieval accuracy, even though CAFe undergoes MMEB-specific fine-tuning following multitask training. These improvements are primarily attributed to the compression-based prompt design and unified training with compression-driven objectives.

\vspace{-0.8em}
\paragraph{Multimodal Language Generation}
As shown in Tab. \ref{tab:mllm}, our method enables the model to achieve state-of-the-art multimodal retrieval performance while largely preserving its generative capabilities. To isolate the effect of generation training data, we directly fine-tune Qwen2-VL on the ShareGPT-4V with the same batch sizes and training steps (\modelname$_{G}$) for reference. Results show that \modelname~achieves performance comparable to both the original and fine-tuned baselines, while models trained only on the retrieval tasks (\modelname$_{R}$) exhibit a significant performance drop, particularly in open-ended question scenarios such as MMVet. This demonstrates the compatibility of our unified framework with generative tasks, without compromising overall model capabilities.

\begin{figure*}[t]
    \centering
    \includegraphics[width=0.85\linewidth]{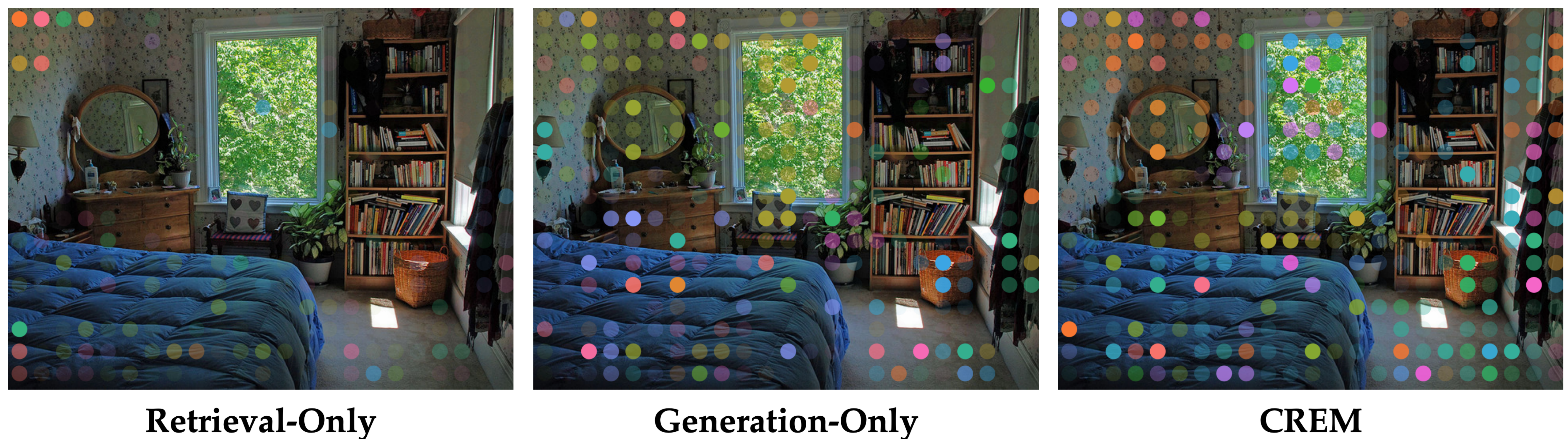}
    \caption{\textbf{Visualization of Chorus Token Attention.} We visualize the attention weights from chorus tokens to vision tokens. Each chorus token is assigned a unique color, and each vision token is colored based on its most attended chorus token, with color intensity reflecting attention strength.}
    \label{fig:crem-visual}
\end{figure*}
\section{Analysis}
\label{sec:analysis}

\paragraph{Analysis of Retrieval and Generation}
We further analyze the relationship between retrieval and generation tasks, as presented in Tab.~\ref{tab:abla-ret-gen}. 
Our baseline experiments, which train a retrieval-specific model on MMEB and a generation-specific model on ShareGPT-4V, reveal a clear limitation: \emph{each model performs well on its primary task but shows severely degraded or even absent capability on the alternate one.} 
To address this, we introduce a \emph{compression-based prompt design} (CPD) and train models separately on retrieval and generation datasets, where the generation task is optimized under a \emph{compression-driven strategy} (CTS). 
The results demonstrate that \textbf{retrieval performance benefits notably from the enlarged optimization space and the use of generation-style prompt templates.} 
At the same time, compression-driven training preserves multimodal understanding while enabling efficient token compression during inference. 
Notably, performance on the retrieval tasks also improves, indicating that introducing compressed representations facilitates cross-task transfer.
We further evaluate a simple mixed-training approach by combining the retrieval objective with generative supervision through an additive loss. 
We observe that retrieval and generation performance degrade when the tasks are naively trained together, even with generation-style templates and chorus tokens.
These findings highlight the importance of the \emph{compression-driven training strategy} (CTS), where \textbf{consistent and informative generative supervision plays a crucial role in strengthening retrieval representations.}

\begin{table}[t]
\setlength{\extrarowheight}{1pt}
\setlength{\aboverulesep}{0.5pt}
\setlength{\belowrulesep}{0.5pt}
\centering
\caption{\textbf{Analysis on Retrieval and Generation.} ``Ret.'' denotes training on retrieval datasets; ``Gen.'' denotes training on generation datasets. ``CPD'' indicates the compression-based prompt design, and ``CTS'' refers to the compression-driven training strategy.}
\label{tab:abla-ret-gen}
\resizebox{\linewidth}{!}{
\begin{tabular}{ccccccc} 
\toprule
\multirow{2}{*}{\textbf{Ret.}} & \multirow{2}{*}{\textbf{Gen.}} & \multirow{2}{*}{\textbf{CPD}} & \multirow{2}{*}{\textbf{CTS}} & \multicolumn{2}{c}{\textbf{Generation}} & \textbf{Retrieval} \\ 
\cline{5-7}
 &  &  &  & Nat. & Comp. & MMEB \\ 
\midrule
\checkmark &  &  &  & 43.4 & - & 62.3 \\
 & \checkmark &  &  & 53.2 & - & 2.9 \\
\checkmark &  & \checkmark &  & 47.2 & - & 66.1 \\
 & \checkmark & \checkmark & \checkmark & 53.0 & 43.9 & 21.1 \\ 
\hdashline
\checkmark & \checkmark & \checkmark &  & 52.8 & - & 65.5 \\
\rowcolor[rgb]{0.91,0.91,0.91} \checkmark & \checkmark & \checkmark & \checkmark & \textbf{53.1} & \textbf{44.2} & \textbf{66.7} \\
\bottomrule
\end{tabular}}
\end{table}

\begin{table}[t]
\setlength{\extrarowheight}{1pt}
\setlength{\aboverulesep}{0.5pt}
\setlength{\belowrulesep}{0.5pt}
\centering
\caption{\textbf{Analysis on Different Chorus Token Design.} ``CTok.'' refers to the type or the number of chorus token. ``Cache'' denotes the KV cache ratio during decoding, with 100\% corresponds to full caching ($\sim$ 1280 tokens). }
\label{tab:number_chorus_tokens}
\begin{tabular}{ccccc} 
\toprule
\textbf{CTok.} & \textbf{MMEB} & \textbf{Nat.} & \textbf{Comp.} & \textbf{Cache(\%)} \\ 
\midrule
\textit{EOS} & 62.3 & 43.4 & - & 100\% \\
\hdashline
1 & 65.6 & 51.9 & 38.2 & 0.07\% \\
4 & 66.1 & 52.3 & 41.7 & 0.28\% \\
8 & 66.5 & 52.5 & 43.1 & 0.56\% \\
\rowcolor[rgb]{0.91,0.91,0.91} \textbf{16} & \textbf{66.7} & \textbf{53.1} & \textbf{44.2} & \textbf{1.12\%} \\
32 & 66.7 & 52.9 & 44.6 & 2.24\% \\
64 & 66.6 & 53.1 & 46.2 & 4.48\% \\
\bottomrule
\end{tabular}
\end{table}

\vspace{-0.8em}
\paragraph{Analysis on Chorus Token}
We examine the effect of the type and number of chorus tokens on retrieval and generation performance. 
Conventional embedding models typically rely on a single token (e.g., the EOS token) for retrieval. 
As shown in Tab.~\ref{tab:number_chorus_tokens}, using the EOS token severely degrades generative capability, while replacing it with an unused special token as chorus token alleviates this issue. 
We further observe that increasing the number of representation tokens leads to moderate gains in retrieval accuracy but eventually results in performance degradation beyond a certain threshold. 
This finding reflects the trade-off between sparse and dense representation spaces. 
Considering both performance and compression efficiency, we set the default number of chorus tokens to 16.

\vspace{-0.8em}
\paragraph{Analysis on Generation Data Mixing}
During joint training of retrieval and generation, we utilize a designed mixture of homogeneous and heterogeneous generation data. As summarized in Tab.~\ref{tab:gen-data-mix}, we further investigate the impact of each data source on model performance. Incorporating homogeneous data under compression-driven joint training leads to notable improvements in generation quality, although it still falls short of original performance. In addition, it contributes to enhanced retrieval embedding quality, demonstrating the synergistic effect of homogeneous examples. On the other hand, training exclusively with heterogeneous data substantially boosts comprehension capabilities, yet yields only modest gains in embedding quality. Striking a balance by combining both data types in a mixed training scheme produces the best overall performance, simultaneously enhancing representation quality and maintaining robust generalization across understanding tasks.

\begin{table}[t]
\setlength{\extrarowheight}{1pt}
\setlength{\aboverulesep}{0.5pt}
\setlength{\belowrulesep}{0.5pt}
\centering
\caption{\textbf{Analysis on Generation Data Mixing.} ``HMD'' denotes training with homogeneous data, which are pseudo-labeled from retrieval pairs, while ``HTD'' denotes training with heterogeneous data collected from open-source datasets.}
\label{tab:gen-data-mix}
\begin{tabular}{ccccc} 
\toprule
\multirow{2}{*}{\textbf{HMD}} & \multirow{2}{*}{\textbf{HTD}} & \multicolumn{2}{c}{\textbf{Generation}} & \textbf{Retrieval} \\ 
\cline{3-5}
 &  & Nat. & Comp. & MMEB \\ 
\midrule
 &  & 47.2 & - & 66.1 \\
\checkmark &  & 49.8 & 42.5 & 66.5 \\
 & \checkmark & 53.0 & 44.0 & 66.2 \\
\rowcolor[rgb]{0.91,0.91,0.91} \checkmark & \checkmark & \textbf{53.1} & \textbf{44.2} & \textbf{66.7} \\
\bottomrule
\end{tabular}
\end{table}

\paragraph{Visualization}
We visualize the spatial focus of chorus tokens under different training paradigms (retrieval-only model, compression-driven generation-only model, and our proposed \modelname). We prompt the model with \textit{"\textless Image\textgreater\textbackslash nRepresent the given image"} and get attention scores from all chorus tokens to vision tokens at the intermediate layer. 
As shown in Fig.~\ref{fig:crem-visual}, retrieval-only models yield sparse attention focused on a few regions, reflecting redundancy. Compression-driven generation leads to broader attention but with limited global coverage. In contrast, \modelname~distributes attention more evenly across the image, capturing distinct and complementary regions.

\section{Conclusion}
\label{sec:conclusion}
In this work, we propose \textbf{\modelname}, which bridges retrieval and comprehension through a unified compression-driven framework. By combining compression-based prompt design with joint optimization of contrastive and generative objectives, our method improves multimodal representations while preserving competitive generative performance. Extensive results on MMEB and multiple comprehension benchmarks demonstrate its effectiveness, offering a scalable path for future unified representation learning and generative modeling.
{
    \small
    \bibliographystyle{ieeenat_fullname}
    \bibliography{main}
}
\appendix
\setcounter{page}{1}

\section{More Implementation Details}

\paragraph{Training Details}
As shown in Tab.~\ref{tab:training-parameters}, we follow most hyperparameter configurations from~\cite{vlm2vec,vlm2vec_v2}. A cosine annealing learning rate schedule is adopted for both retrieval and generation tasks. We apply LoRA with rank 16 and  alpha 64, targeting the projection layers of query, key, value, and output. This preserves the model's comprehension capacity without degrading retrieval performance. Images are processed using dynamic resolution and MRoPE, with the number of image tokens constrained between 256 and 1280. Training is conducted for 2000 steps with a 100-step warmup. All experiments are run on 8 NVIDIA A800 GPUs. Details of the retrieval training datasets are listed in Tab.~\ref{tab:mmeb_statistics}.

\begin{table}[ht!]
\centering
\caption{Training Hyperparameters and Computational Requirements for Retrieval and Generation Tasks.}
\label{tab:training-parameters}
\begin{tabular}{lcc} 
\toprule
\textbf{Hyperparameter} & Retrieval & Generation \\ 
\midrule
Training Samples & 1.1M & 665K \\
Batch Size & 1024 & 128 \\
Resize Tokens & \multicolumn{2}{c}{(256, 1280)} \\
Learning rate & \multicolumn{2}{c}{$5 \times 10^{-5}$} \\
Optimizer & \multicolumn{2}{c}{AdamW} \\
Learning Rate Decay & \multicolumn{2}{c}{cosine} \\
Loss Weight & 1 & 0.5 \\
Warmup Steps & \multicolumn{2}{c}{100} \\
LoRA Rank & \multicolumn{2}{c}{16} \\
LoRA Alpha & \multicolumn{2}{c}{64} \\
LoRA Target Modules & \multicolumn{2}{c}{$q,k,v,o$} \\
Temperature $\tau$ & \multicolumn{2}{c}{0.02} \\
Training Steps & \multicolumn{2}{c}{2000} \\
GPU Configuration & \multicolumn{2}{c}{8$\times$A800} \\
\bottomrule
\end{tabular}
\end{table}

\begin{table*}[!ht]
\setlength{\extrarowheight}{0.5pt}
\small
\centering
\caption{\textbf{Zero-shot cross-domain retrieval performance.} Results are reported as average Recall@1 across short-caption, long-caption, and compositional retrieval benchmarks. The best results are in bold.}
\vspace{-2mm}
\label{tab:unime}
\resizebox{\textwidth}{!}{
 \begin{tabular}{lccccccccccc}
\toprule
 \multicolumn{1}{c}{\multirow{4}{*}{\textbf{Models}}} & \multicolumn{4}{c}{\textbf{Short Caption Retrieval}} & \multicolumn{4}{c}{\textbf{Long Caption Retrieval}} & \multicolumn{3}{c}{\textbf{Compositional Retrieval}} \\
\cmidrule(lr){2-5} \cmidrule(lr){6-9} \cmidrule(lr){10-12}
 & \multicolumn{2}{c}{Flickr30K} & \multicolumn{2}{c}{COCO} & \multicolumn{2}{c}{ShareGPT4V} & \multicolumn{2}{c}{Urban1K} & \multicolumn{3}{c}{SugarCrepe} \\
\cmidrule(lr){2-3} \cmidrule(lr){4-5} \cmidrule(lr){6-7} \cmidrule(lr){8-9} \cmidrule(lr){10-12}
 & $q^t\rightarrow c^i$ & $q^i\rightarrow c^t$ & $q^t\rightarrow c^i$ & $q^i\rightarrow c^t$ & $q^t\rightarrow c^i$ & $q^i\rightarrow c^t$ & $q^t\rightarrow c^i$ & $q^i\rightarrow c^t$  & Replace & Swap & Add \\
\midrule
VLM2Vec~\cite{vlm2vec} & 76.0 & 90.6 & 46.8  & 66.6 & 85.8 & 90.7 & 84.7 & 90.8 & 85.8 & 66.3 & 86.5 \\
UniME~\cite{unime} & \textbf{83.3} & 94.4 & 54.8 & 74.0 & \textbf{93.9} & 89.3 & 94.3 & 95.5 & 80.5 & 65.5 & 82.2 \\
\hdashline
\rowcolor[rgb]{0.91,0.91,0.91}
\textbf{\modelname~2B} & 81.4 & 91.9 & 55.8 & 71.6 & 90.3 & 89.4 & 94.0 & 94.3 & 87.9 & 66.1 & 91.5 \\
\rowcolor[rgb]{0.91,0.91,0.91}
\textbf{\modelname~7B} & 82.9 & \textbf{95.5} & \textbf{56.4} & \textbf{76.1} & 90.7 & \textbf{90.8} & \textbf{97.6} & \textbf{98.0} & \textbf{88.7} & \textbf{74.0} & \textbf{91.7} \\
\bottomrule
\end{tabular}}
\vspace{-3mm}
\end{table*}

\paragraph{Evaluation Details}
For retrieval evaluation, we use the MMEB test set shown in Tab.~\ref{tab:mmeb_statistics}. Generation tasks are evaluated based on VLMEvalKit~\cite{duan2024vlmevalkit}, which supports over 80 benchmarks. We report MMBench~\cite{liu2024mmbench} results based on its English version, as our model is trained on corpora only in English.

\section{More Results and Analysis}

\paragraph{Detailed Results on MMEB}

Per-task results on MMEB across 36 tasks are presented in Tab.~\ref{tab:app_mmeb_per_task}. Some potentially stronger baselines are excluded due to incomplete score reporting. \modelname~2B and \modelname~7B variants shown are trained with high-resolution images, using up to 1280 vision tokens as in UNITE~\cite{kong2025unite}.

\paragraph{Cross-domain Evaluations on More Benchmarks}
Following the evaluation protocol in UniME~\cite{unime}, we further validate the cross-domain generalization of our model across three distinct dimensions: short-caption retrieval, long-caption retrieval, and compositional reasoning. As summarized in Tab.~\ref{tab:unime}, our model, particularly the \modelname-7B variant, demonstrates superior cross-domain generalization, outperforming LLaVA-OV-based UniME in most metrics. This performance edge is particularly evident in complex long-form text matching and fine-grained compositional understanding, highlighting the effectiveness of our compression-driven representation in capturing and preserving intricate multimodal semantics.

\paragraph{Ablations on Training Hyperparameters}
As shown in Tab.~\ref{tab:abla-loss-weight}, we conduct an ablation study on the loss weights for retrieval ($\alpha_r$) and generation ($\alpha_g$), as well as the probability $p$ used in stochastic compression-driven language modeling. We observe that reducing the generation loss weight yields only a marginal impact on generative quality while slightly improving embedding performance. Based on these observations, we set $\alpha_r = 1$ and $\alpha_g = 0.5$ to achieve a balanced trade-off between retrieval accuracy and comprehension ability. Furthermore, applying compression-driven training for all samples weakens native generative capability and provides only limited gains in retrieval performance, whereas pure generative training does not enhance representation quality, lacks compression ability, and leads to inferior retrieval results. Therefore, we set $p = 0.5$ to balance generative fidelity and retrieval effectiveness.

\begin{table}[!ht]
\setlength{\extrarowheight}{0.5pt}
\setlength{\aboverulesep}{0.5pt}
\setlength{\belowrulesep}{0.5pt}
\centering
\caption{\textbf{Ablation on Loss Weights and Compression Probability.} $\alpha_r$ and $\alpha_g$ denote the loss weights for retrieval and generation objectives, respectively, while $p$ represents the probability of applying compression-driven generation.}
\label{tab:abla-loss-weight}
\begin{tabular}{cccccc} 
\toprule
\multirow{2}{*}{\textbf{$\alpha_r$}} & \multirow{2}{*}{\textbf{$\alpha_g$}} & \multirow{2}{*}{\textbf{$p$}} & \multicolumn{2}{c}{\textbf{Generation}} & \textbf{Retrieval} \\ 
\cline{4-6}
 &  &  & Nat. & Comp. & MMEB \\ 
\midrule
1 & 1 & 1 & 52.4 & 45.3 & 66.4 \\
1 & 1 & 0.5 & 53.1 & 44.5 & 66.3 \\
1 & 1 & 0 & 52.9 & - & 65.3 \\
\rowcolor[rgb]{0.91,0.91,0.91} \textbf{1} & \textbf{0.5} & \textbf{0.5} & \textbf{53.1} & \textbf{44.2} & \textbf{66.7} \\
0.5 & 1 & 0.5 & 53.2 & 45.0 & 65.5 \\
\bottomrule
\end{tabular}
\end{table}

\paragraph{Ablations on Pooling Methods}

As shown in Tab.~\ref{tab:pooling_method}, we evaluate several pooling strategies for aggregating information from chorus tokens into a single retrieval embedding. Using a mlp network followed by mean pooling degrades performance, and attention pooling with a learnable query token also underperforms compared to simple mean pooling. We attribute this to weak supervision signals during contrastive training, which may hinder the pooling module from effectively learning to extract informative features across tokens.

\begin{table}[t]
\setlength{\extrarowheight}{0.5pt}
\setlength{\aboverulesep}{0.5pt}
\setlength{\belowrulesep}{0.5pt}
\centering
\caption{\textbf{Ablation on Pooling Methods.} Evaluation of various pooling strategies used to aggregate chorus tokens for retrieval. ``CLS'': classification, ``QA'': question answering, ``RET'': retrieval, ``GD'': grounding.}
\label{tab:pooling_method}
\begin{tabular}{lccccc} 
\toprule
\textbf{Pooling} & CLS & QA & RET & GD & Overall \\ 
\midrule
MLP & 64.6 & 60.7 & 67.6 & 76.0 & 65.8 \\
Attention & 63.6 & 59.4 & 67.7 & 76.7 & 65.3 \\
\rowcolor[rgb]{0.91,0.91,0.91}\textbf{Mean} & \textbf{65.8} & \textbf{60.7} & \textbf{68.3} & \textbf{78.9} & \textbf{66.7} \\
\bottomrule
\end{tabular}
\end{table}

\section{Limitations}

Although our framework achieves impressive performance on both retrieval and generation, several limitations remain. The fixed number of chorus tokens after training limits adaptability, as optimal compression may vary between tasks. Dynamic or task-aware token allocation is a promising direction for future work. Compression-based inference leads to poor performance in OCR tasks, likely due to impaired fine-grained visual understanding, which could be alleviated by incorporating more OCR-specific data. Additionally, our training is primarily based on MMEB and ShareGPT-4V, and incorporating more diverse, high-quality retrieval and generation data may further improve generalization.

\begin{table*}[ht!]
\setlength{\extrarowheight}{1pt}
\setlength{\aboverulesep}{0.5pt}
\setlength{\belowrulesep}{0.5pt}
\centering
\caption{\textbf{Details of Retrieval Data.} MMEB consists of 36 datasets across four meta-task categories. Of these, 20 in-distribution datasets are used for training and 16 out-of-distribution datasets are reserved for evaluation.}
\label{tab:mmeb_statistics}
\begin{tabular}{c|ccccccc}
\toprule
Meta-Task & Dataset & Query$\to$Target & Distribution Type & \#Training & \#Eval & \#Candidates \\ 
\midrule
\multirow{10}{*}{\begin{tabular}[c]{@{}c@{}}Classification \\ (10 Tasks)\end{tabular}} & ImageNet-1K       & I$\to$T         & IND & 100K & 1000 & 1000 \\  
& N24News           & I + T$\to$I     & IND &  49K & 1000 & 24   \\ 
& HatefulMemes      & I$\to$T         & IND &   8K & 1000 & 2    \\ 
& VOC2007           & I$\to$T         & IND &   8K & 1000 & 20   \\ 
& SUN397            & I$\to$T         & IND &  20K & 1000 & 397  \\ 
\cmidrule(lr){2-7}
& Place365          & I$\to$T         & OOD & -    & 1000 & 365  \\ 
& ImageNet-A        & I$\to$T         & OOD & -    & 1000 & 1000 \\ 
& ImageNet-R        & I$\to$T         & OOD & -    & 1000 & 200  \\ 
& ObjectNet         & I$\to$T         & OOD & -    & 1000 & 313  \\ 
& Country-211       & I$\to$T         & OOD & -    & 1000 & 211  \\
\midrule
\multirow{10}{*}{\begin{tabular}[c]{@{}c@{}}VQA \\ (10 Tasks)\end{tabular}} 
& OK-VQA            & I + T$\to$T     & IND &9K&1000 &1000 \\  
& A-OKVQA           & I + T$\to$T     & IND &  17K & 1000 & 1000 \\ 
& DocVQA            & I + T$\to$T     & IND &  40K & 1000 & 1000 \\ 
& InfographicVQA    & I + T$\to$T     & IND &  24K & 1000 & 1000 \\ 
& ChartQA           & I + T$\to$T     & IND &  28K & 1000 & 1000 \\  
& Visual7W          & I + T$\to$T     & IND &  70K & 1000 & 1000 \\ 
\cmidrule(lr){2-7}
& ScienceQA         & I + T$\to$T     & OOD & -   & 1000 & 1000 \\  
& VizWiz            & I + T$\to$T     & OOD & -   & 1000 & 1000 \\ 
& GQA               & I + T$\to$T     & OOD & -   & 1000 & 1000 \\ 
& TextVQA           & I + T$\to$T     & OOD & -   & 1000 & 1000 \\
\midrule
\multirow{12}{*}{\begin{tabular}[c]{@{}c@{}}Retrieval \\ (12 Tasks)\end{tabular}} 
& VisDial           & T$\to$I         & IND & 123K & 1000 & 1000 \\ 
& CIRR              & I + T$\to$I     & IND &  26K & 1000 & 1000 \\ 
& VisualNews\_t2i   & T$\to$I         & IND & 100K & 1000 & 1000 \\ 
& VisualNews\_i2t   & I$\to$T         & IND & 100K & 1000 & 1000 \\ 
& MSCOCO\_t2i       & T$\to$I         & IND & 100K & 1000 & 1000 \\ 
& MSCOCO\_i2t       & I$\to$T         & IND & 113K & 1000 & 1000 \\ 
& NIGHTS            & I$\to$I         & IND &  16K & 1000 & 1000 \\ 
& WebQA             & T$\to$I + T     & IND &  17K & 1000 & 1000 \\ 
\cmidrule(lr){2-7}
& OVEN              & I + T$\to$I + T & OOD & -   & 1000 & 1000 \\ 
& FashionIQ         & I + T$\to$I     & OOD & -   & 1000 & 1000 \\ 
& EDIS              & T$\to$I + T     & OOD & -   & 1000 & 1000 \\ 
& Wiki-SS-NQ        & T $\to$I        & OOD & -   & 1000 & 1000 \\
\midrule
\multirow{4}{*}{\begin{tabular}[c]{@{}c@{}}Visual Grounding\\ (4 Tasks)\end{tabular}} 
& MSCOCO            & I + T$\to$I     & IND & 100K & 1000 & 1000 \\  
\cmidrule(lr){2-7}
& Visual7W-Pointing & I + T$\to$I     & OOD & -   & 1000 & 1000 \\ 
& RefCOCO           & I + T$\to$I     & OOD & -   & 1000 & 1000 \\ 
& RefCOCO-Matching  & I + T$\to$I + T & OOD & -   & 1000 & 1000 \\
\bottomrule
\end{tabular}
\end{table*}

\begin{table*}[ht]
\centering
\setlength{\extrarowheight}{1pt}
\setlength{\aboverulesep}{0.5pt}
\setlength{\belowrulesep}{0.5pt}
\caption{\textbf{Detailed MMEB Results.} Performance of baseline models and our \modelname~across 20 in-distribution (IND) and 16 out-of-distribution (OOD) datasets. OOD datasets are highlighted with a yellow background. For each baseline, we report the strongest variant with complete evaluation metrics: VLM2Vec 7B (LLaVA-1.6), MMRet 7B (LLaVA-1.6), UniME 7B (LLaVA-1.6), mmE5 11B (Llama-3.2-Vision), and UNITE 7B (Qwen2-VL).}
\label{tab:app_mmeb_per_task}
\scalebox{0.95}{
\begin{tabular}{lcccccccc} 
\toprule
\rowcolor[rgb]{0.851,0.851,0.851}  & \textbf{CLIP} & \textbf{VLM2Vec} & \textbf{MMRet} & \textbf{UniME} & \textbf{mmE5} & \textbf{UNITE} & \textbf{\modelname~2B} & \textbf{\modelname~7B} \\ 
\midrule
\rowcolor[rgb]{1,0.851,0.702} \textbf{Classification (10 tasks)} &  &  &  &  &  &  &  &  \\
ImageNet-1K & 55.8 & 74.5 & 58.8 & 71.3 & 77.8 & 80.2 & 82.6 & 83.6 \\
N24News & 34.7 & 80.3 & 71.3 & 79.5 & 81.7 & 80.3 & 77.8 & 81.2 \\
HatefulMemes & 51.1 & 67.9 & 53.7 & 64.6 & 64.2 & 67.1 & 62.1 & 65.0 \\
VOC2007 & 50.7 & 91.5 & 85.0 & 90.4 & 91.0 & 84.9 & 80.8 & 85.6 \\
SUN397 & 43.4 & 75.8 & 70.0 & 75.9 & 77.7 & 78.7 & 77.5 & 77.3 \\
\rowcolor[rgb]{1,0.992,0.851} Place365 & 28.5 & 44.0 & 43.0 & 45.6 & 43 & 44.5 & 41.5 & 44.0 \\
\rowcolor[rgb]{1,0.992,0.851} ImageNet-A & 25.5 & 43.6 & 36.1 & 45.5 & 56.3 & 59.2 & 47.6 & 54.5 \\
\rowcolor[rgb]{1,0.992,0.851} ImageNet-R & 75.6 & 79.8 & 71.6 & 78.4 & 86.3 & 90.5 & 90.8 & 90.2 \\
\rowcolor[rgb]{1,0.992,0.851} ObjectNet & 43.4 & 39.6 & 55.8 & 36.4 & 62.5 & 68.1 & 72.2 & 71.4 \\
\rowcolor[rgb]{1,0.992,0.851} Country-211 & 19.2 & 14.7 & 14.7 & 18.7 & 35.4 & 29.5 & 25.1 & 30.1 \\
\textit{All Classification} & 42.8 & 61.2 & 56.0 & 60.6 & 67.6 & 68.3 & 65.8 & 68.3 \\ 
\midrule
\rowcolor[rgb]{0.702,0.702,1} \textbf{VQA (10 tasks)} &  &  &  &  &  &  &  &  \\
OK-VQA & 7.5 & 69.0 & 73.3 & 68.3 & 67.6 & 67.1 & 63.9 & 71.8 \\
A-OKVQA & 3.8 & 54.4 & 56.7 & 58.7 & 56.1 & 58.0 & 53.4 & 61.4 \\
DocVQA & 4.0 & 52.0 & 78.5 & 67.6 & 90.3 & 92.7 & 91.7 & 94.2 \\
InfographicsVQA & 4.6 & 30.7 & 39.3 & 37.0 & 56.5 & 71.3 & 65.2 & 76.2 \\
ChartQA & 1.4 & 34.8 & 41.7 & 33.4 & 50.5 & 63.2 & 53.3 & 67.2 \\
Visual7W & 4.0 & 49.8 & 49.5 & 51.7 & 51.9 & 54.9 & 52.7 & 57.4 \\
\rowcolor[rgb]{1,0.992,0.851} ScienceQA & 9.4 & 42.1 & 45.2 & 40.5 & 55.8 & 51.2 & 42.5 & 59.2 \\
\rowcolor[rgb]{1,0.992,0.851} VizWiz & 8.2 & 43.0 & 51.7 & 42.7 & 52.8 & 53.4 & 48.2 & 53.6 \\
\rowcolor[rgb]{1,0.992,0.851} GQA & 41.3 & 61.2 & 59.0 & 63.6 & 61.7 & 56.8 & 54.0 & 66.5 \\
\rowcolor[rgb]{1,0.992,0.851} TextVQA & 7.0 & 62.0 & 79.0 & 65.2 & 83.3 & 82.3 & 82.4 & 86.4 \\
\textit{All VQA} & 9.1 & 49.9 & 57.4 & 52.9 & 62.6 & 65.1 & 60.7 & 69.4 \\ 
\midrule
\rowcolor[rgb]{0.702,1,0.702} \textbf{Retrieval (12 tasks)} &  &  &  &  &  &  &  &  \\
VisDial & 30.7 & 80.9 & 83.0 & 79.7 & 74.1 & 80.5 & 80.5 & 85.6 \\
CIRR & 12.6 & 49.9 & 61.4 & 52.2 & 54.7 & 51.6 & 58.2 & 62.5 \\
VisualNews\_t2i & 78.9 & 75.4 & 74.2 & 74.8 & 77.6 & 79.3 & 71.3 & 79.5 \\
VisualNews\_i2t & 79.6 & 80.0 & 78.1 & 78.8 & 83.3 & 82.4 & 76.9 & 83.2 \\
MSCOCO\_t2i & 59.5 & 75.7 & 78.6 & 74.9 & 76.4 & 78.2 & 75.1 & 78.3 \\
MSCOCO\_i2t & 57.7 & 73.1 & 72.4 & 73.8 & 73.2 & 74.3 & 72.2 & 73.8 \\
NIGHTS & 60.4 & 65.5 & 68.3 & 66.2 & 68.3 & 66.0 & 66.8 & 67.6 \\
WebQA & 67.5 & 87.6 & 90.2 & 89.8 & 88.0 & 87.0 & 89.3 & 90.8 \\
\rowcolor[rgb]{1,0.992,0.851} FashionIQ & 11.4 & 16.2 & 54.9 & 16.5 & 28.8 & 26.3 & 16.1 & 23.3 \\
\rowcolor[rgb]{1,0.992,0.851} Wiki-SS-NQ & 55.0 & 60.2 & 24.9 & 66.6 & 65.8 & 72.2 & 61.0 & 73.2 \\
\rowcolor[rgb]{1,0.992,0.851} OVEN & 41.1 & 56.5 & 87.5 & 55.7 & 77.5 & 73.1 & 67.4 & 73.1 \\
\rowcolor[rgb]{1,0.992,0.851} EDIS & 81.0 & 87.8 & 65.6 & 86.2 & 83.7 & 88.3 & 84.5 & 84.4 \\
\textit{All Retrieval} & 53.0 & 67.4 & 69.9 & 67.9 & 71.0 & 71.6 & 68.3 & 72.9 \\ 
\midrule
\rowcolor[rgb]{0.925,0.702,0.776} \textbf{Visual Grounding (4 tasks)} &  &  &  &  &  &  &  &  \\
MSCOCO & 33.8 & 80.6 & 76.8 & 76.5 & 53.7 & 73.9 & 65.0 & 69.5 \\
\rowcolor[rgb]{1,0.992,0.851} RefCOCO & 56.9 & 88.7 & 89.8 & 89.3 & 92.7 & 89.2 & 86.5 & 92.0 \\
\rowcolor[rgb]{1,0.992,0.851} RefCOCO-matching & 61.3 & 84.0 & 90.6 & 90.6 & 88.8 & 90.1 & 89.3 & 93.6 \\
\rowcolor[rgb]{1,0.992,0.851} Visual7W-pointing & 55.1 & 90.9 & 77.0 & 84.1 & 92.3 & 86.1 & 74.7 & 89.3 \\
\textit{All Visual Grounding} & 51.8 & 86.1 & 83.6 & 85.1 & 89.6 & 84.8 & 78.9 & 86.1 \\ 
\midrule
\rowcolor[rgb]{0.851,0.953,0.992} \textbf{Final Score (36 tasks)} &  &  &  &  &  &  &  &  \\
All & 37.8 & 62.9 & 64.1 & 66.6 & 69.8 & 70.3 & 66.7 & 72.1 \\
All IND & 37.1 & 67.5 & 59.1 & 68.4 & 72.3 & 73.6 & 70.8 & 75.6 \\
All OOD & 38.7 & 57.1 & 68.0 & 57.9 & 66.7 & 66.3 & 61.5 & 67.8 \\
\bottomrule
\end{tabular}}
\end{table*}


\end{document}